\documentclass[runningheads]{llncs}

\usepackage{eccv}

\usepackage{eccvabbrv}

\usepackage[dvipsnames,table]{xcolor}
\usepackage{graphicx}
\usepackage{booktabs}
\usepackage{multirow}
\usepackage{tikz-imagelabels}
\usepackage{graphicx}
\usepackage{caption}
\usepackage{subcaption}
\definecolor{darkerblue}{rgb}{0,0.08,0.45}
\usepackage{wrapfig}
\usepackage{adjustbox}
\usepackage{makecell}

\definecolor{color4}{rgb}{0.94,0.94,1}
\definecolor{CB}{rgb}{0.21, 0.49, 0.74}

\usepackage[accsupp]{axessibility}  

\usepackage[pagebackref,breaklinks,colorlinks,citecolor=eccvblue]{hyperref}

\usepackage{orcidlink}

\begin{document}

\title{Fast and Accurate Image Restoration and Generation with Rank Enhanced Linear Attention} 

\titlerunning{Fast and Accurate Image Restoration and Generation with Linear Attention}

\author{Yuang Ai\orcidlink{0009-0005-6445-898X}}

\authorrunning{Y.~Ai}

\institute{The Chinese University of Hong Kong \\ \email{yuang@link.cuhk.edu.hk}}

\maketitle

\begin{abstract}
Transformer-based models have made remarkable progress in image restoration (IR) tasks. However, the quadratic complexity of self-attention in Transformer hinders its applicability to high-resolution images. Existing methods mitigate this issue with sparse or window-based attention, yet inherently limit global context modeling. Linear attention, a variant of softmax attention, demonstrates promise in global context modeling while maintaining linear complexity, offering a potential solution to the above challenge. Despite its efficiency benefits, vanilla linear attention suffers from a significant performance drop in IR, largely due to the low-rank nature of its attention map. To counter this, we propose Rank Enhanced Linear Attention (RELA), a simple yet effective method that enriches feature representations by integrating a lightweight depthwise convolution. Building upon RELA, we propose an efficient and effective Vision Transformer, named LAformer. LAformer eliminates hardware-inefficient operations such as softmax and window shifting, enabling efficient processing of high-resolution images. Extensive experiments across 7 IR tasks and 21 benchmarks demonstrate that LAformer outperforms SOTA methods and offers significant computational advantages. Furthermore, we extend LAformer to diffusion-based and flow-based visual generation, showcasing its strong potential as a competitive alternative to DiT and SiT. Code and models are available at \url{https://github.com/shallowdream204/LAformer}.
\end{abstract}

\section{Introduction}
\label{sec:intro}
Image restoration (IR) aims to recover a high-quality (HQ) image by removing degradations from its degraded low-quality (LQ) counterpart. 
As a fundamental task in low-level vision, image restoration has a wide array of real-world applications, spanning fields such as autonomous driving~\cite{zhu2016traffic}, surveillance monitoring~\cite{rasti2016convolutional}, medical imaging~\cite{greenspan2009super}, \etc. Consequently, developing hardware-efficient IR networks holds substantial practical value.

Over the past decade, convolutional neural networks (CNNs) have achieved remarkable progress in various image restoration tasks~\cite{dong2015image,zhang2017beyond,xu2014deep}. However, CNNs are often limited by their constrained receptive field, making it challenging to capture long-range dependencies.
Recently, Transformer-based methods~\cite{dat,restormer,ai2024dreamclear}
have significantly advanced the performance of image restoration.

\begin{figure}[!t]
    \centering
    \includegraphics[width=0.8\linewidth]{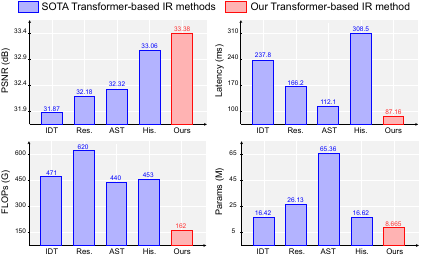}
    \caption{Comprehensive comparison with SOTA Transformer-based methods for image restoration (raindrop removal).}
    \label{fig:psnr}
\end{figure}

While Transformer offers the advantage of a global receptive field, its core self-attention mechanism introduces quadratic computational complexity. Directly applying self-attention globally incurs substantial overhead, particularly when handling high-resolution images. To mitigate this, some methods compute self-attention within local windows~\cite{swinir,uformer,chen2023activating}, while others adopt sparse attention patterns~\cite{zhang2022accurate,drsformer}. While effectively reducing computational costs, these methods either limit global perception or are sensitive to degradation types due to specific attention patterns~\cite{ast}.

\begin{figure}[!t]
    \centering
    \includegraphics[width=0.8\linewidth]{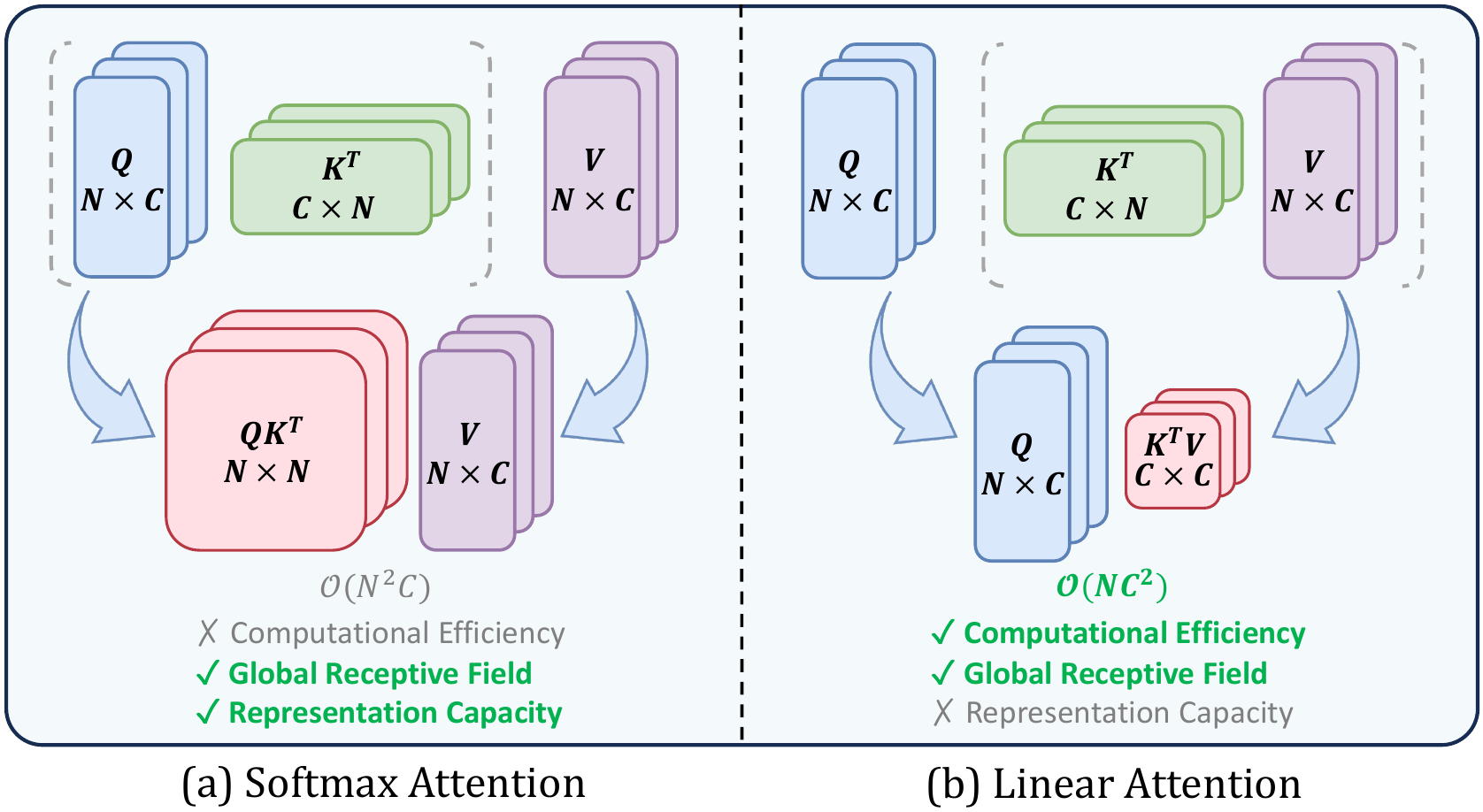}
    \caption{Conceptual comparison between \textbf{Softmax Attention} and \textbf{Linear Attention}. While linear attention excels at efficiently capturing global context, its representation capacity is limited, resulting in a notable performance drop.}
    \label{fig:fig1}
\end{figure}

The above analysis leads to a crucial question: \textit{Is it possible to design a hardware-efficient image restoration model that simultaneously preserves strong global perceptive capabilities?}
In this paper, we explore linear attention~\cite{katharopoulos2020transformers} as a potential solution to this issue. Compared to standard softmax attention~\cite{attention}, linear attention replaces the softmax operation with simpler activation functions (\eg, ReLU~\cite{relu}, ELU~\cite{elu}) and reorders the computation to first calculate $K^{\top} V$. As shown in Fig.~\ref{fig:fig1}, linear attention reduces the computational complexity to a linear scale with respect to the number of visual tokens $N$, while preserving the global contextual advantages of softmax attention. When processing high-resolution images, the global receptive field of linear attention enables the capture of long-range relationships between pixels, thereby enhancing restoration quality. These attributes motivate us to employ linear attention in developing an efficient and effective image restoration model.

However, standard linear attention mechanisms appear suboptimal for image restoration, resulting in a notable performance drop relative to softmax attention (see Tab.~\ref{tab:ablation_model}).
Through rank analysis~\cite{grl}, we find that the low-rank nature of the attention map in linear attention severely limits the diversity of its output features, thereby diminishing its representation capacity (see Fig.~\ref{fig:rank}).
To overcome this limitation, we propose Rank Enhanced Linear Attention (RELA), which incorporates a lightweight depthwise convolution to effectively enhance the rank of output feature. RELA retains linear complexity while boosting the representation capacity of linear attention, enabling improved modeling of complex degradation patterns.

\begin{figure}[t] 
\centering
    \includegraphics[width=1.0\textwidth]{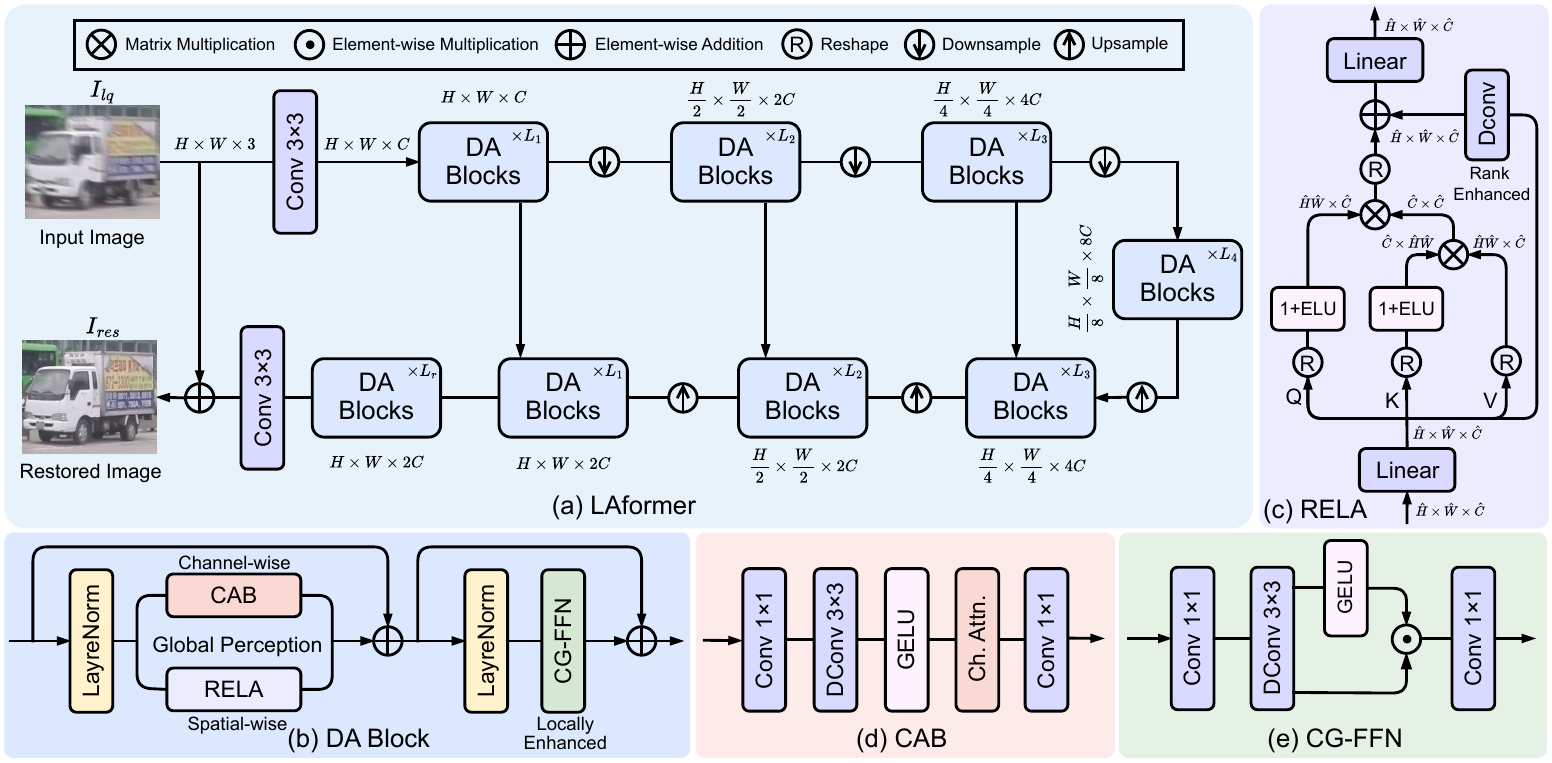}
    \caption{\textbf{Overall architecture of LAformer,} which includes (b) Dual Attention (DA) Block, (c) Rank Enhanced Linear Attention (RELA), (d) Channel Attention Block (CAB), and (e) Convolutional-Gated Feed-Forward Network (CG-FFN).} 
    \label{fig:arch}
\end{figure}

We further propose LAformer, an efficient Vision Transformer structured within the widely adopted U-Net architecture~\cite{restormer,diffir}. As shown in Fig.~\ref{fig:arch}, LAformer incorporates a Dual-Attention (DA) Block that operates across spatial and channel dimensions, leveraging both RELA and channel attention~\cite{zhang2018image} concurrently to facilitate efficient global information capture.
Recognizing the crucial role of local information in restoring texture details, we introduce a Convolutional Gated Feed-Forward Network (CG-FFN) to strengthen LAformer’s capability in modeling spatial locality. This architecture enables LAformer to effectively integrate local and global information, enhancing its representational power.
Notably, LAformer circumvents hardware-inefficient operations like softmax and window shifting, significantly boosting its efficiency for high-resolution image processing. As shown in Fig.~\ref{fig:psnr}, our LAformer surpasses SOTA Transformer-based IR methods and significantly reduces computational overhead. Furthermore, we extend LAformer for diffusion and flow-based visual generation, demonstrating its remarkable efficiency and scalability.

Our main contributions can be summarized as follows:
\begin{itemize}
    \item We propose LAformer, a simple, efficient, and strong Vision Transformer that combines powerful representation capability with linear complexity.
    \item We propose Rank Enhanced Linear Attention (RELA), a simple yet effective mechanism that boosts feature diversity in linear attention by enhancing the rank of features.
    \item Extensive experiments across 7 IR tasks and 21 benchmarks validate that LAformer achieves SOTA performance while maintaining high efficiency.
    \item When extended to visual generation, LAformer demonstrates impressive efficiency and scalability, highlighting its potential as a competitive alternative to DiT and SiT.
\end{itemize}

\section{Related Work}
\textbf{Image Restoration.}
Images captured in real-world scenarios often suffer from various degradations, such as blur, low-light, and adverse weather conditions. These issues inevitably degrade the performance of downstream models and diminish the effectiveness of many vision systems~\cite{zhu2016traffic, mao2017can,ai2024uncertainty,ai2026uncertainty}. SRCNN~\cite{dong2015image} and DnCNN~\cite{zhang2017beyond} are pioneering works that introduce CNN into image restoration and achieve impressive results. Subsequently, a plethora of CNN-based architectures incorporating meticulously designed modules have been introduced to enhance IR performance. Notable examples include the residual block~\cite{kim2016accurate,lim2017enhanced,ledig2017photo}, dense block~\cite{zhang2018residual,wang2018esrgan,yulun_res_den}, channel attention~\cite{zhang2018image}, and non-local attention mechanisms~\cite{mei2020image,mei2021image,xia2022efficient}. Despite their success, CNNs remain constrained by their inherently limited receptive field, posing challenges in effectively capturing long-range dependencies.
Recently, researchers have begun to use self-attention mechanisms in place of convolutions to enhance IR performance~\cite{chen2021pre,swinir,restormer,uformer}.

\noindent\textbf{Vision Transformer.}
Since its introduction by NLP researchers, Transformer~\cite{attention} has been widely adopted in computer vision tasks~\cite{dosovitskiy2020image,liu2021swin,Fan_2025_ICCV,fan2026random,ai2026bitdance}. IPT~\cite{chen2021pre} is the pioneering work that introduces Transformer to image restoration by processing images in small patches. Since then, many works~\cite{restormer,grl,zhang2024hit,zhang2023ntire,cao2023ntire} have focused on improving self-attention, aiming to reduce its significant computational overhead. SwinIR~\cite{swinir} and Uformer~\cite{uformer} perform self-attention in non-overlapping local windows. HAT~\cite{chen2023activating} proposes overlapping cross-attention to establish cross-window interaction. ART~\cite{zhang2022accurate} introduces sparse attention to IR for a large receptive field. AST~\cite{ast} proposes an adaptive sparse self-attention to filter out irrelevant tokens. Although effective, a trade-off between global receptive field and computational efficiency persists.

Due to its efficiency and ability to capture global context, Linear attention~\cite{katharopoulos2020transformers} has garnered significant interest from the computer vision community. Shen~\etal\cite{shen2021efficient} introduce linear attention to object detection and instance segmentation. 
Bolya~\etal\cite{bolya2022hydra} propose hydra attention by maximizing the number of linear attention heads. Han~\etal\cite{han2023flatten} propose a focused linear attention module to enhance efficiency and expressiveness. EfficientViT~\cite{cai2023efficientvit} finds that linear attention cannot produce sharp attention distributions and proposes multi-scale linear attention to enhance local information extraction. Distinct from prior works, ours is the first to demonstrate the efficacy of linear attention in high-resolution image restoration. Additionally, we investigate the limitation of linear attention in representation capacity and propose novel approaches to overcome the challenge.

\section{Method}
We propose LAformer, an efficient framework for high-resolution IR. First, we analyze the softmax attention and linear attention mechanisms in Sec.~\ref{sec:preliminary}. Then Sec.~\ref{sec:rela} delves into the inherent low-rank characteristics of linear attention, motivating the development of Rank Enhanced Linear Attention (RELA) as a solution. Finally, in Sec.~\ref{sec:laformer}, we present LAformer, a model built on RELA that achieves explicit global modeling with linear complexity.

\subsection{Softmax vs. Linear Attention}
\label{sec:preliminary}
Softmax attention~\cite{attention} is a widely used mechanism in Transformer. For a given input $X\in\mathbb{R}^{N\times C}$,  where
$N$ represents the number of visual tokens and $C$ denotes the channel dimension, it can be generally formulated as
\begin{equation}
\label{eq:softmax}
    Y_i=\sum_{j=1}^{N} \frac{\mathrm{Sim}(Q_i,K_j) }{\sum_{j=1}^{N}\mathrm{Sim}(Q_i,K_j)} V_j,
\end{equation}
where $Q=XW_Q, K=XW_K, V=XW_V$, $W_{Q,K,V}\in\mathbb{R}^{C\times C}$ are learnable linear projection matrices, $\mathrm{Sim}(\cdot,\cdot)$ is the similarity function, $i,j$ index visual tokens. When the similarity function is set as $\mathrm{Sim}(Q_i,K_j)=\mathrm{exp}(\frac{Q_iK_j^{\top}}{\sqrt{C}})$, Eq.~(\ref{eq:softmax}) represents the standard softmax attention. Due to the need to compute similarities between all queries and keys, softmax attention incurs a $\mathcal{O}(N^2)$ complexity.

Linear attention~\cite{katharopoulos2020transformers} uses simple activation functions to approximate the similarity function as $\mathrm{Sim}(Q_i,K_j)=\psi(Q_i)\psi(K_j)^{\top}$, where $\psi(\cdot)$ is the activation function. With this assumption, we can rewrite  Eq.~(\ref{eq:softmax}) as
\begin{equation}
\label{eq:linear}
\begin{split}
Y_i &= \sum_{j=1}^{N}\frac{\psi(Q_i)\psi(K_j)^{\top}}{\sum_{j=1}^{N}\psi(Q_i)\psi(K_j)^{\top} }V_j \\
&= \frac{\psi(Q_i)\left(\sum_{j=1}^{N}\psi (K_j)^{\top} V_j\right)}{\psi (Q_i)\left(\sum _{j=1}^N\psi (K_j)^{\top}\right)}.
\end{split}
\end{equation}

Compared to softmax attention, linear attention modifies the computation order from $(QK^{\top})V$ to $Q(K^{\top}V)$, reducing the complexity from $\mathcal{O}(N^2)$ to $\mathcal{O}(N)$ while preserving the global modeling capabilities. However, the improvement in computational efficiency is accompanied by a noticeable drop in performance (see Tab.~\ref{tab:ablation_model}).

\begin{figure}[!t]
    \centering
    \includegraphics[width=0.9\linewidth]{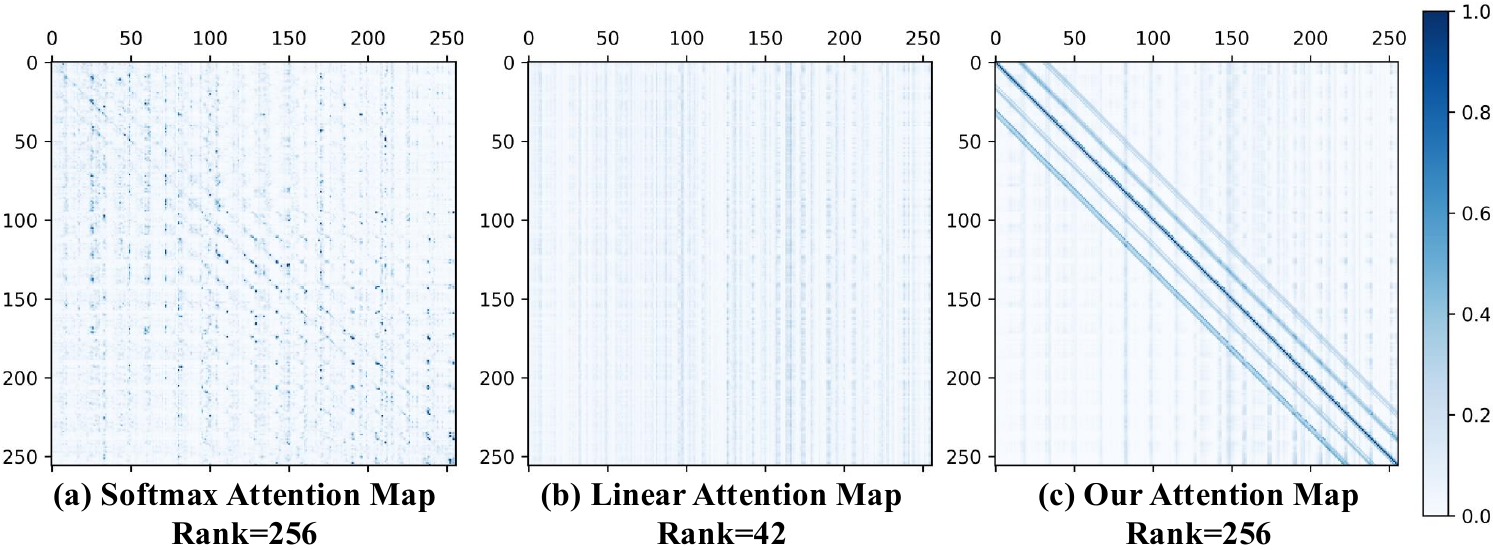}
    \caption{\textbf{Visualization of attention maps} ($N=256, C=48$) of softmax attention (window-based), linear attention, and RELA. Our RELA effectively restores the rank of the attention map.
    }
    \label{fig:rank_attn}
\end{figure}

\subsection{Rank Enhanced Linear Attention}
\label{sec:rela}
Following~\cite{grl}, we conduct rank analysis for the linear attention map $M$, which is computed as $\psi(Q)\psi(K)^{\top}\in \mathbb{R}^{N\times N}$. According to basic matrix theory, it follows that
\begin{equation}
\begin{split}
    \mathrm{Rank}(M) &\le \mathrm{min}( \mathrm{Rank}(\psi(Q)),
    \mathrm{Rank}(\psi(K)^{\top})\\ &\le \mathrm{min}(N,C).
\end{split}
\end{equation}

\begin{figure}[!t]
    \centering
    \includegraphics[width=0.9\linewidth]{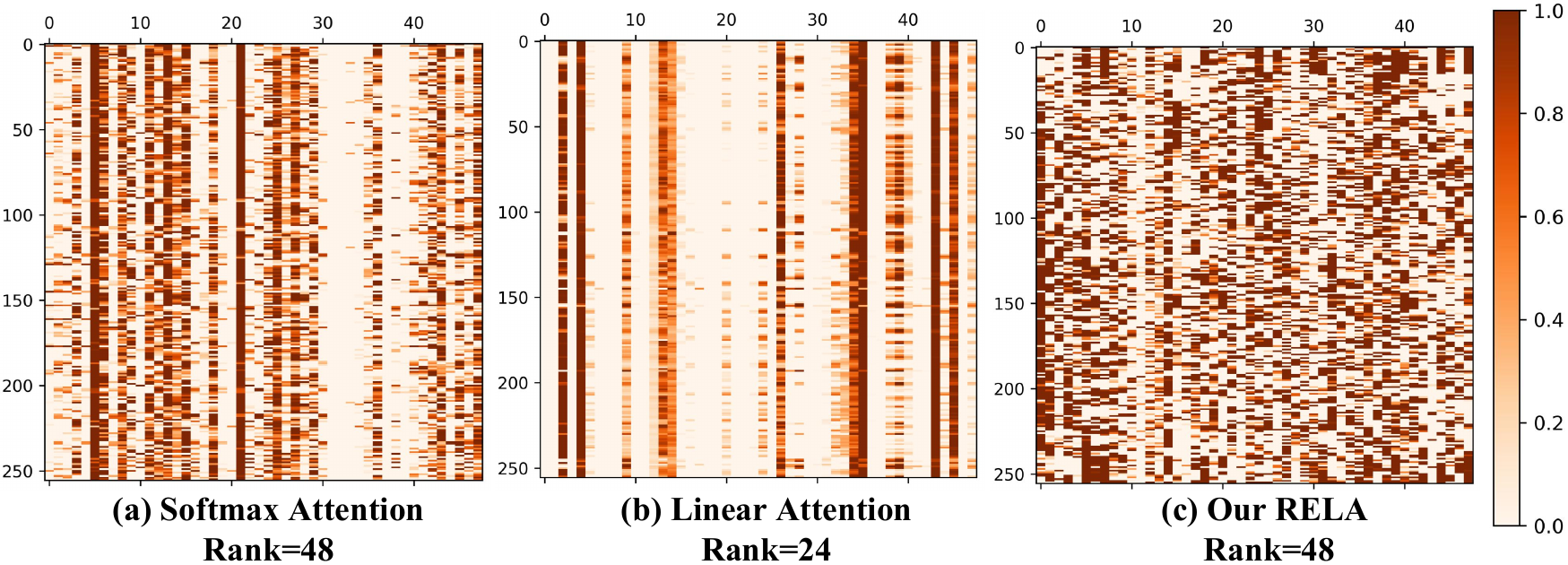}
    \caption{\textbf{Visualization of feature maps} output by softmax attention (window-based), linear attention, and RELA. Our RELA substantially enriches the expressiveness of the resulting features.}
    \label{fig:rank}
\end{figure}

For high-resolution IR, the number of visual tokens $N$ is typically much larger than $C$, which results in the low-rank nature of the attention map in linear attention (see Fig.~\ref{fig:rank_attn}). Since the attention output is a weighted sum of $V$ using the attention map, the highly correlated weights inevitably limit the expressive capacity of the output features. As shown in Fig.~\ref{fig:rank}, the rank of output features produced by linear attention is notably lower than that of softmax attention, suggesting a reduced diversity in features.

To address this limitation, we introduce Rank Enhanced Linear Attention (RELA), depicted in Fig.~\ref{fig:arch} (c). We integrate a lightweight depth-wise convolution (DWC) into the computation process, which can be formulated as
\begin{equation}
    Y = (1+\mathrm{ELU} (Q))(1+\mathrm{ELU} (K))^{\top}V+W_dV,
    \label{eq:rela}
\end{equation}
where $W_d$ represents the depth-wise convolution. Following~\cite{katharopoulos2020transformers}, we use $1+\mathrm{ELU}(\cdot)$ as the activation function.

The DWC can be interpreted as a local attention operation, enriching the output through local feature combinations. We can rewrite Eq.~(\ref{eq:rela}) as $Y=(\psi(Q)\psi(K)^{\top}+M_{DWC})V=M_{attn}V$,
where $M_{DWC}$ represents the sparse matrix derived from the DWC operation, and $M_{attn}$ denotes the resulting full attention matrix. Since $M_{DWC}$ can potentially be of full rank, its incorporation effectively raises the upper bound of the rank of the overall attention map $M_{attn}$. Fig.~\ref{fig:rank_attn} and Fig.~\ref{fig:rank} illustrate that our RELA successfully restores the attention map and output features to a full-rank state, substantially enhancing the diversity of learned features. It is worth noting that our RELA introduces only a minimal increase in parameters and computational overhead compared to standard linear attention, yet it brings a significant performance boost (see Tab.~\ref{tab:ablation_model}).

\subsection{Network Architecture}
\label{sec:laformer}
\subsubsection{Overall Architecture}
The overall architecture of our LAformer is depicted in Fig.~\ref{fig:arch}. To rigorously assess the effectiveness of our proposed method, we adopt the classical U-Net architecture as the backbone, aligning with previous approaches~\cite{restormer,uformer,nafnet}. For an input low-quality (LQ) image $I_{lq}\in \mathbb{R}^{H\times W\times 3}$, LAformer initially applies a $3\times 3$ convolution for overlapping patch embedding, transforming $I_{lq}$ into a feature map with $C$ channels.  In both the encoder and decoder modules, we incorporate Dual Attention (DA) Blocks to efficiently capture both global and local degradation patterns.
Within each stage, skip connections link the encoder and decoder, enabling seamless information flow across intermediate features. Across stages, pixel-unshuffle and pixel-shuffle operations are employed for effective feature down-sampling and up-sampling. 

\subsubsection{Dual Attention Block}
To improve both global and local feature extraction in LAformer, we integrate a dual attention module and a Convolutional Gated Feed-Forward Network (CG-FFN), which address global and local feature learning, respectively. The details of these components are outlined below.

\noindent\textbf{Dual Attention.} We use the proposed RELA to achieve efficient global modeling in the spatial dimension. Considering that channel-wise information also plays an important role in IR, we introduce channel attention~\cite{zhang2018image} into the DA Block, forming the Channel Attention Block (CAB) in conjunction with convolutions, as depicted in Fig.~\ref{fig:arch} (d). The process of CAB can be formulated as
\begin{equation}
    Y=W_{p_2}\mathrm{CA}(\mathrm{GELU}(W_dW_{p_1}X)),
\end{equation}
where $W_{p_{(\cdot)}}$ is the $1\times1$ point-wise convolution, $W_d$ is the $3\times3$ depth-wise convolution, $\mathrm{CA}(\cdot)$ denotes the channel attention operation. Channel attention can be interpreted as a mechanism for modeling global dependencies, as it aggregates token information across the spatial domain through global average pooling.
In the DA Block, we combine RELA and CAB to effectively model and aggregate global information across both spatial and channel dimensions, achieving efficient global modeling and aggregation.

\noindent\textbf{Convolutional Gated FFN.} As discussed in~\cite{cai2023efficientvit}, compared to softmax attention, linear attention tends to produce less sharp attention maps, which hinders its ability to effectively capture local information. However, rich local information is crucial for restoring texture details. To address this, we propose a simple Convolutional Gated Feed-Forward Network (CG-FFN) to enhance the local fitting capability of LAformer. As illustrated in Fig.~\ref{fig:arch} (e), CG-FFN integrates depth-wise convolution with Gated Linear Unit (GLU)~\cite{dauphin2017language}, which can be formulated as
\begin{equation}
    Y=W_{p_2}(\mathrm{GELU}( W_dW_{p_1}X)\odot W_dW_{p_1}X).
\end{equation}
Compared to MLP, CG-FFN incorporates depth-wise convolution to enhance LAformer's capacity for capturing fine-grained local details while reducing computational cost, thereby complementing the linear attention mechanism.

\noindent\textbf{Dual-Attention Block.} Our DA Block is equipped with RELA, CAB, and CG-FFN, which is defined as
\begin{equation}
\begin{split}
    X^\prime &= \mathrm{RELA}(\mathrm{LN}(X)) + \mathrm{CAB}(\mathrm{LN}(X)) + X, \\
    Y &= \mathrm{CGFFN}(\mathrm{LN}(X^\prime)) + X^\prime.
\end{split}
\end{equation}

Built upon the stacked DA Blocks, LAformer avoids hardware-inefficient operations (\eg, softmax, window shifting) and explicitly models spatial-wise global features, channel-wise global features, and local features, all with linear complexity. This enables LAformer to maintain high efficiency while possessing strong representation capability.

\begin{table}[b]
\scriptsize
\center
\begin{center}
\caption{{\textbf{Motion deblurring}} results on real-world datasets. All models are trained and tested on RealBlur~\cite{realblur}.}
\label{tab:motion_real}
\scalebox{0.62}{
\setlength{\tabcolsep}{3pt}
        \begin{tabular}{c c c c c c c c c c c c >{\columncolor{color4}}c}
        \toprule[0.15em]
         &  & DeblurGAN-v2 & SRN & MIMO-UNet+ & MPRNet & DeepRFT+ & BANet & MSSNet & Stripformer & FFTformer & GRL & \textbf{LAformer-B} \\
        \multirow{-2}{*}{\textbf{Dataset}} & \multirow{-2}{*}{\textbf{Method}} & \cite{kupyn2019deblurgan} & \cite{tao2018scale} & \cite{cho2021rethinking} & \cite{zamir2021multi} & \cite{tsai2022banet}& \cite{kim2022mssnet}  & \cite{mao2021deep} & \cite{stripformer} &\cite{kong2023efficient}& \cite{grl} & \textbf{Ours} \\
        \midrule[0.15em]
        \textbf{RealBlur-R} & PSNR & 36.44 & 38.65 & N/A & 39.31 & 39.55 & 39.76 &  39.84& 39.84 & 40.11 & \underline{40.20} & \textbf{41.07}\\ 
        \cite{realblur} & SSIM & 0.935 & 0.965 & N/A & 0.972 & 0.971 & 0.972& 0.972&0.974 &0.973& \underline{0.974} & \textbf{0.977}\\
        \midrule
        \textbf{RealBlur-J} & PSNR & 29.69 & 31.38 & 31.92 & 31.76 & 32.00 &32.10 &32.19& 32.48 &32.62&\underline{32.82} &\textbf{32.92}\\
        \cite{realblur} & SSIM & 0.870 & 0.909 & 0.919 & 0.922 & 0.923 & 0.928 &0.931& 0.929 &\underline{0.933}&0.932& \textbf{0.933}\\ 
        \bottomrule[0.15em]
        \end{tabular}
}
\end{center}
\end{table}

\section{Experiments}
\subsection{Experimental Setup}
We evaluate the proposed LAformer on various datasets for \textbf{7 image restoration tasks}: \textbf{(1)} motion deblurring, \textbf{(2)} defocus deblurring, \textbf{(3)} image dehazing, \textbf{(4)} image desnowing, \textbf{(5)} raindrop removal, \textbf{(6)} low-light enhancement, and \textbf{(7)} all-in-one image restoration. Furthermore, we extend LAformer for \textbf{diffusion-based and flow-based visual generation}. Additional details about the training dataset, training protocols, and more visual results can be found in the \textbf{Appendix} due to space limits.

\noindent \textbf{Implementation Details.} The kernel size for the depth-wise convolution in RELA is set to $5 \times 5$. To align with SOTA models for each specific task, we adapt the number of blocks and channels, considering variants of different sizes, including Tiny, Small, and Base versions (LAformer-T/S/B). All experiments are conducted on A100 GPUs. Additional task-specific training hyperparameters and details of the model architecture can be found in the \textbf{Appendix}.

\begin{figure*}[!t]
	
	\centering
	
	\newcommand{\h}{0.105}
	\newcommand{\wa}{0.12}
	\newcommand{\wb}{0.16}
	\newcommand{\g}{-0.7mm}
	\setlength\tabcolsep{1.8pt}
  	\renewcommand{\arraystretch}{1}
	\resizebox{1.00\linewidth}{!} {
			\renewcommand{\h}{0.143}
			\newcommand{\w}{0.22}
				\begin{adjustbox}{valign=t}
					\begin{tabular}{ccccccc}
						\includegraphics[height=\h \textwidth, width=\w \textwidth]{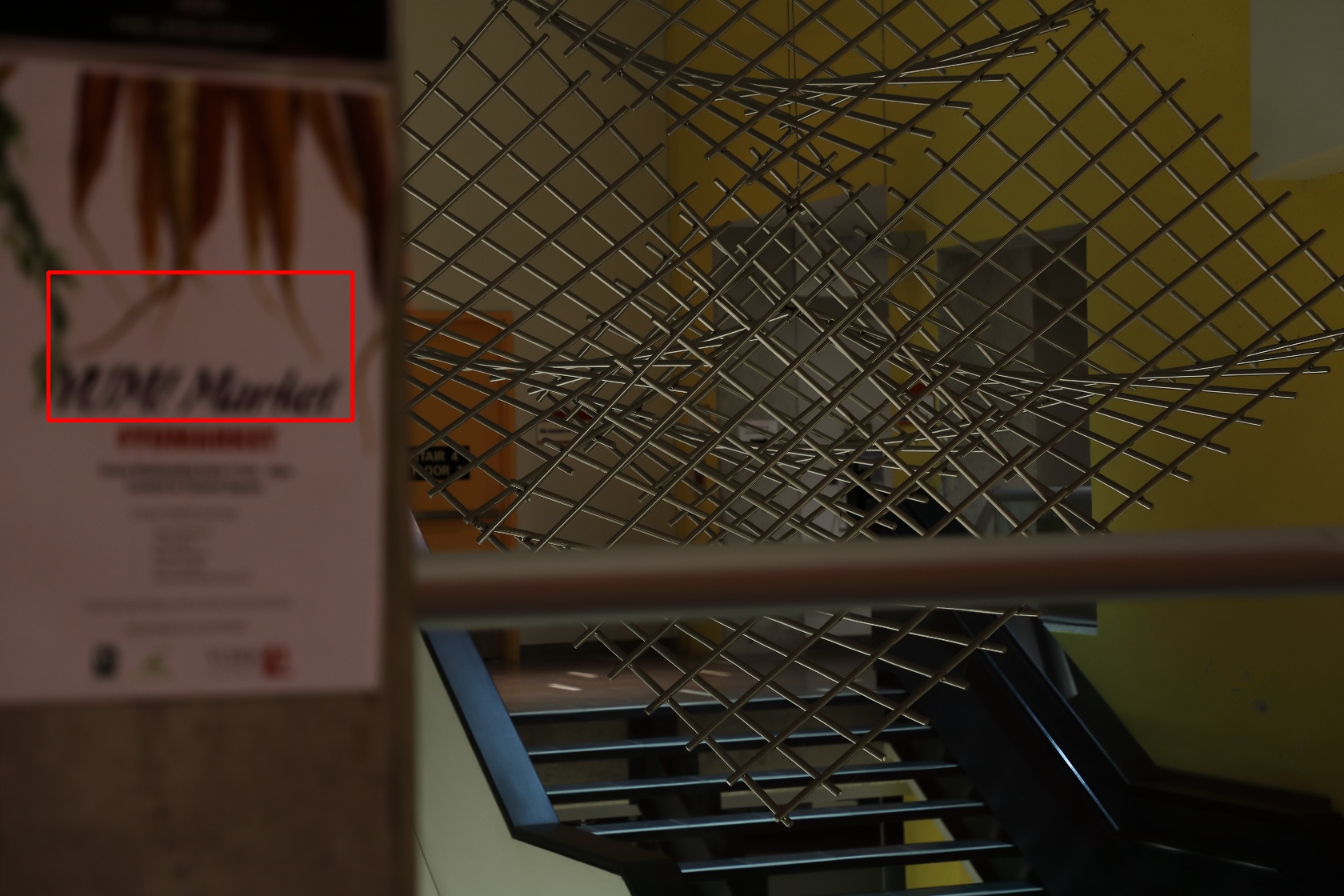} \hspace{\g} &
						\includegraphics[height=\h \textwidth, width=\w \textwidth]{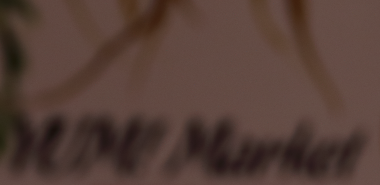} \hspace{\g} &
				\includegraphics[height=\h \textwidth, width=\w \textwidth]{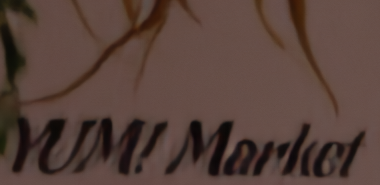} \hspace{\g} &
      			\includegraphics[height=\h \textwidth, width=\w \textwidth]{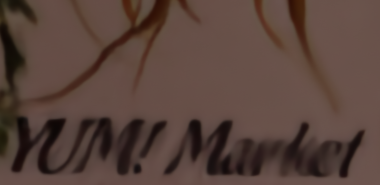} \hspace{\g} &
      			\includegraphics[height=\h \textwidth, width=\w \textwidth]{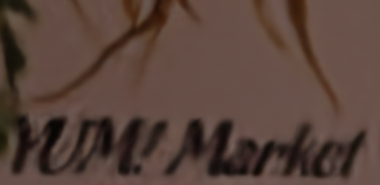} \hspace{\g} &
      			\includegraphics[height=\h \textwidth, width=\w \textwidth]{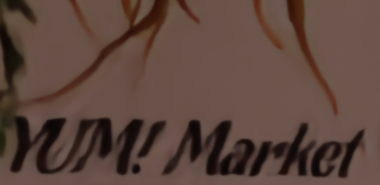} \hspace{\g} &
				\includegraphics[height=\h \textwidth, width=\w \textwidth]{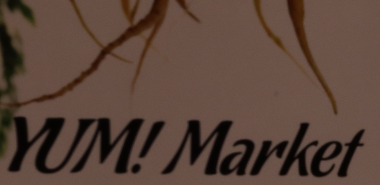} 
						\\
						Blurry Image \hspace{\g} &
						Input \hspace{\g} &
						DRBNet~\cite{drbnet} \hspace{\g} &
						Restormer~\cite{restormer} \hspace{\g} &
						NRKNet~\cite{nrknet} \hspace{\g} &
      					\textbf{LAformer} \hspace{\g} &
                            Target \hspace{\g}
						\\
      					PSNR/SSIM \hspace{\g} &
						23.29/0.731 \hspace{\g} &
						25.12/0.796 \hspace{\g} &
						27.03/0.856 \hspace{\g} &
						25.57/0.798\hspace{\g} &
      					\textbf{30.22}/\textbf{0.916} \hspace{\g} &
                            $+\infty$/1.000 \hspace{\g}
						\\
					\end{tabular}
				\end{adjustbox}
			
	}
   	\renewcommand{\arraystretch}{1}
	\resizebox{1.00\linewidth}{!} {
			\renewcommand{\h}{0.143}
			\newcommand{\w}{0.22}
				\begin{adjustbox}{valign=t}
					\begin{tabular}{ccccccc}
						\includegraphics[height=\h \textwidth, width=\w \textwidth]{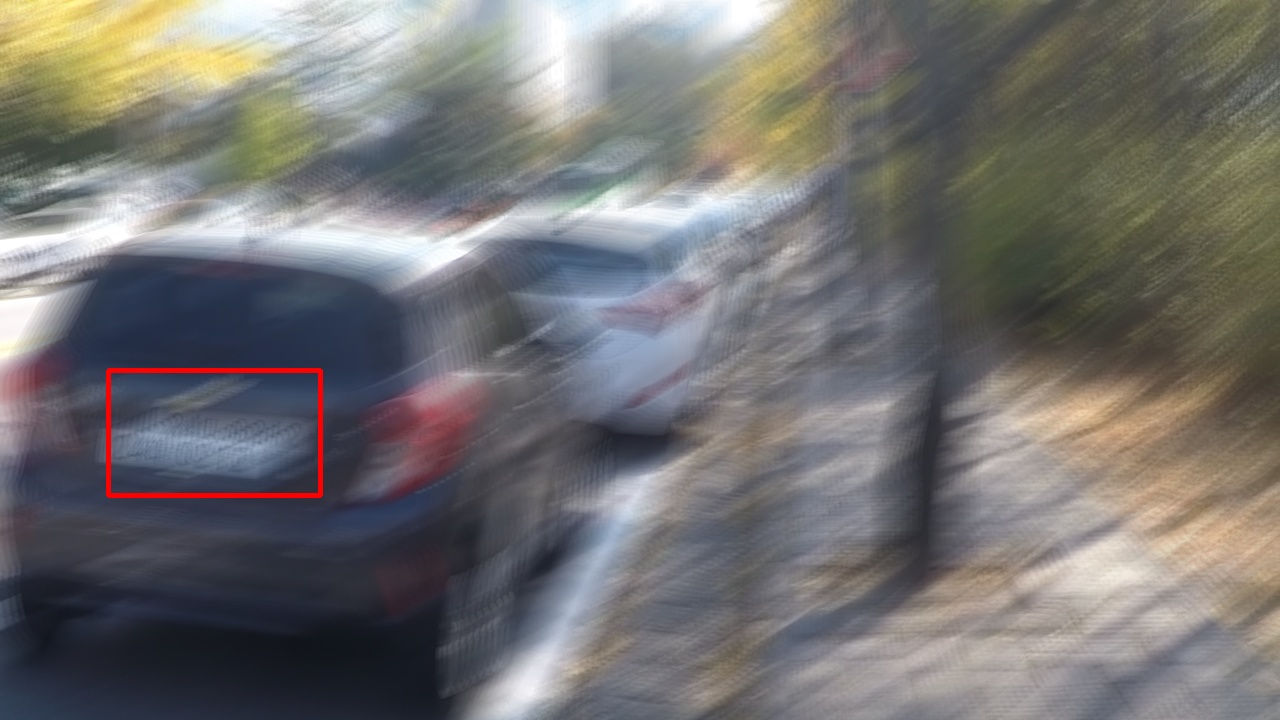} \hspace{\g} &
						\includegraphics[height=\h \textwidth, width=\w \textwidth]{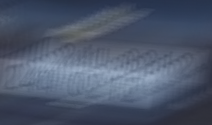} \hspace{\g} &
				\includegraphics[height=\h \textwidth, width=\w \textwidth]{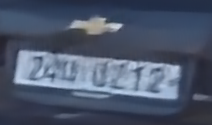} \hspace{\g} &
      			\includegraphics[height=\h \textwidth, width=\w \textwidth]{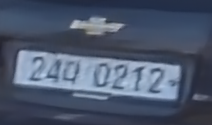} \hspace{\g} &
      			\includegraphics[height=\h \textwidth, width=\w \textwidth]{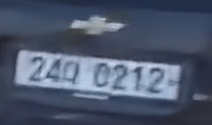} \hspace{\g} &
      			\includegraphics[height=\h \textwidth, width=\w \textwidth]{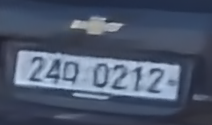} \hspace{\g} &
				\includegraphics[height=\h \textwidth, width=\w \textwidth]{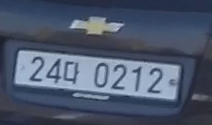} 
						\\
						Blurry Image \hspace{\g} &
						Input \hspace{\g} &
						MPRNet~\cite{zamir2021multi} \hspace{\g} &
						Restormer~\cite{restormer} \hspace{\g} &
						FPro~\cite{fpro} \hspace{\g} &
      					\textbf{LAformer} \hspace{\g} &
                            Target \hspace{\g}
						\\
      					PSNR/SSIM \hspace{\g} &
						18.99/0.526 \hspace{\g} &
						25.22/0.794 \hspace{\g} &
						28.39/0.879 \hspace{\g} &
						27.28/0.853 \hspace{\g} &
      					\textbf{28.91}/\textbf{0.895} \hspace{\g} &
                            $+\infty$/1.000 \hspace{\g}
						\\
					\end{tabular}
				\end{adjustbox}
			
	}
	\caption{Top row: Visual comparison on the DPDD~\cite{dpdd} dataset for {\textbf{single-image defocus deblurring}}. Bottom row: Visual comparison on the GoPro~\cite{gopro} dataset for {\textbf{motion deblurring}}. Zoom in for a better view.}
	\label{fig:visual_defocus_motion}
\end{figure*}

\begin{table}[t]
    \centering
    \begin{minipage}[t]{0.47\linewidth}
        \centering
        \setlength{\tabcolsep}{1.5pt} 
                \caption{\textbf{Motion deblurring} results on synthetic datasets.}
        \resizebox{\linewidth}{!}{
            \begin{tabular}{lccccc} 
            \toprule[0.15em]
            \multirow{2}{*}{\textbf{Method}} & \multirow{2}{*}{\textbf{Venue}} & \multicolumn{2}{c}{\textbf{GoPro}~\cite{gopro}} & \multicolumn{2}{c}{\textbf{HIDE}~\cite{shen2019human}} \\ 
            \cmidrule(lr){3-4} \cmidrule(lr){5-6}
             & & PSNR $\uparrow$ & SSIM $\uparrow$ & PSNR $\uparrow$ & SSIM $\uparrow$ \\ \midrule
            MIMO-Unet+~\cite{cho2021rethinking} & ICCV'21 & 32.45 & 0.957 & 29.99 & 0.930 \\
            IPT~\cite{chen2021pre} & CVPR'21 & 32.52 & N/A & N/A & N/A \\
            MPRNet~\cite{zamir2021multi}& CVPR'21 & 32.66 & 0.959 & 30.96 & 0.939 \\
            Restormer~\cite{wang2023promptrestorer}& CVPR'22 & 32.92 & 0.961 & 31.22 & 0.942 \\
            Stripformer~\cite{stripformer} & ECCV'22 & 33.08& 0.962& 31.03& 0.940 \\
            HI-Diff~\cite{hidiff} & NIPS'23 & \underline{33.33} & \underline{0.964} & \underline{31.46} & \underline{0.945} \\
            IRNeXt~\cite{cui2023irnext}& ICML'23 & 33.16 & 0.962 & N/A & N/A \\
            PromptRestorer~\cite{wang2023promptrestorer} & NIPS'23 & 33.06 & 0.962 & 31.36 & 0.944 \\
            FPro~\cite{fpro} & ECCV'24 & 33.05 & 0.961 & N/A & N/A \\
            ACL~\cite{Gu_2025_CVPR} & CVPR'25 & 33.25 & 0.964 & N/A & N/A \\
            \rowcolor{color4}
            \textbf{LAformer-B} & - &\textbf{33.40} & \textbf{0.965} & \textbf{31.49} & \textbf{0.945} \\ \bottomrule[0.15em]
            \end{tabular}
        }
        \label{tab:motion}
    \end{minipage}
    \hfill 
    \begin{minipage}[t]{0.501\linewidth}
        \centering
        \setlength{\tabcolsep}{1.5pt}
                \caption{\textbf{Image desnowing} results on three widely used datasets.}

        \resizebox{\linewidth}{!}{
            \begin{tabular}{l|cc|cc|cc}
            \toprule[0.15em]
            \multirow{2}{*}{\textbf{Method}} &
            \multicolumn{2}{c|}{\textbf{CSD}~\cite{csd}} & \multicolumn{2}{c|}{\textbf{SRRS}~\cite{srrs}} & \multicolumn{2}{c}{\textbf{Snow100K}~\cite{snow100k}} \\         
             & PSNR & SSIM & PSNR & SSIM & PSNR & SSIM  \\ \midrule
            DesnowNet~\cite{snow100k} & 20.13 & 0.81 & 20.38 & 0.84 & 30.50 & 0.94 \\
            CycleGAN~\cite{engin2018cycle} & 20.98 & 0.80 & 20.21 & 0.74 & 26.81 & 0.89\\
            All in One~\cite{as2020} & 26.31 & 0.87 & 24.98 & 0.88 & 26.07 & 0.88 \\
            JSTASR~\cite{srrs} & 27.96 & 0.88 & 25.82 & 0.89 & 23.12 & 0.86 \\
            HDCW-Net~\cite{csd} & 29.06 & 0.91 & 27.78 & 0.92 & 31.54 & 0.95 \\
            TransWeather~\cite{valanarasu2022transweather} & 31.76 & 0.93 & 28.29 & 0.92 & 31.82 & 0.93 \\
            NAFNet~\cite{nafnet} & 33.13 & 0.96 & 29.72 & 0.94 & 32.41 & 0.95 \\
            FocalNet~\cite{focalnet} & \underline{37.18} & \underline{0.99} & \underline{31.34} & \underline{0.98} & \underline{33.53} & \underline{0.95} \\
            \rowcolor{color4}
            \textbf{LAformer-T}&\textbf{37.42}  & \textbf{0.99} & \textbf{32.24} & \textbf{0.98} & \textbf{34.24} & \textbf{0.96}  \\
            \bottomrule[0.15em]
            \end{tabular}
        }
        \label{tab:snow}
    \end{minipage}
\end{table}

\begin{table}[t]
\scriptsize
\footnotesize
\center
\caption{\textbf{Model efficiency} comparison with Transformer-based methods on the GoPro~\cite{gopro} dataset ($1280\times720$ resolution).}
\begin{center}
\begin{adjustbox}{width=0.7\textwidth}
\setlength{\tabcolsep}{5pt}    
        \begin{tabular}{c|c c c c>{\columncolor{color4}}c}
        \toprule[0.15em]
         \multirow{2}{*}{\textbf{Method}} &Uformer & Restormer & Stripformer & HI-Diff& \textbf{LAformer-B} \\
        &\cite{uformer} & \cite{restormer} & \cite{stripformer} & \cite{hidiff} & \textbf{Ours} \\
        \midrule[0.15em]
        \#Params (M) &50.88& 26.13& 19.71&28.49& 24.89 \\ 
        FLOPs (G) &86.89& 154.88&155.03& 142.62& 144.33 \\ 
        Runtime (s) &1.41& 0.79&0.87& 1.56& 0.73 \\ 
        PSNR (dB) &32.97& 32.96&33.08& 33.33& 33.40 \\ 
        \bottomrule[0.15em]
        \end{tabular}
\end{adjustbox}
\end{center}
\label{tab:efficiency}
\end{table}

\begin{table*}[t]
\begin{center}
\setlength{\tabcolsep}{2pt}
\caption{{\textbf{Defocus deblurring}} results on the DPDD testset~\cite{dpdd}. \textbf{S:} single-image defocus deblurring. \textbf{D:} dual-pixel defocus deblurring.}
\scalebox{0.68}{
\begin{tabular}{l | c c c c | c c c c | c c c c }
\toprule[0.15em]
  \multirow{2}{*}{\textbf{Method}} & \multicolumn{4}{c|}{\textbf{Indoor Scenes}} & \multicolumn{4}{c|}{\textbf{Outdoor Scenes}} & \multicolumn{4}{c}{\textbf{Combined}} \\
\cline{2-13}
  & PSNR~$\textcolor{black}{\uparrow}$ & SSIM~$\textcolor{black}{\uparrow}$& MAE~$\textcolor{black}{\downarrow}$ & LPIPS~$\textcolor{black}{\downarrow}$  & PSNR~$\textcolor{black}{\uparrow}$ & SSIM~$\textcolor{black}{\uparrow}$& MAE~$\textcolor{black}{\downarrow}$ & LPIPS~$\textcolor{black}{\downarrow}$  & PSNR~$\textcolor{black}{\uparrow}$ & SSIM~$\textcolor{black}{\uparrow}$& MAE~$\textcolor{black}{\downarrow}$ & LPIPS~$\textcolor{black}{\downarrow}$   \\
\midrule[0.15em]
DRBNet$_S$~\cite{drbnet}& N/A & N/A &N/A &N/A &N/A &N/A &N/A &N/A &{25.73} & 0.791 & N/A & 0.183 \\
Restormer$_S$~\cite{restormer}& \underline{28.87}  & \underline{0.882}  & \underline{0.025} & \textbf{0.145} & \underline{23.24}  & \underline{0.743}  & \underline{0.050} & \underline{0.209}  & {25.98}  & \underline{0.811}  & \underline{0.038}  & \underline{0.178}   \\
NRKNet$_S$~\cite{nrknet}& N/A & N/A &N/A &N/A &N/A &N/A &N/A &N/A &\underline{26.11} & 0.810 & N/A & 0.210 \\
INIKNet$_S$~\cite{quan2023single}& N/A & N/A &N/A &N/A &N/A &N/A &N/A &N/A &26.05& 0.803& N/A & 0.185 \\
MPerceiver$_S$~\cite{ai2024multimodal}& N/A & N/A &N/A &N/A &N/A &N/A &N/A &N/A &25.88& 0.803& N/A & 0.190 \\
\rowcolor{color4}
\textbf{LAformer-B}$_S$  &  \textbf{29.13} & \textbf{0.886} & \textbf{0.024} & \underline{0.146} &  \textbf{23.57}& \textbf{0.759} & \textbf{0.048} & \textbf{0.207} & \textbf{26.28} & \textbf{0.821} & \textbf{0.036} & \textbf{0.177} \\
\midrule[0.15em]
\midrule[0.15em]
RDPD$_D$~\cite{abdullah2021rdpd} & 28.10 & 0.843 & 0.027 & 0.210 & 22.82 & 0.704 & 0.053 & 0.298 & 25.39 & 0.772 & 0.040 & 0.255 \\

Uformer$_D$~\cite{uformer} & 28.23 & 0.860 & 0.026 & 0.199 & 23.10 & 0.728 & 0.051 & 0.285 & 25.65 & 0.795 & 0.039 & 0.243 \\

IFAN$_D$~\cite{lee2021iterative} & {28.66} & {0.868} & {0.025} & {0.172} & {23.46} & {0.743} & {0.049} & {0.240} & {25.99} & {0.804} & {0.037} & {0.207} \\

Restormer$_D$~\cite{restormer}& \underline{29.48}  & \underline{0.895}  & \underline{0.023} & \underline{0.134} & \underline{23.97}  & \underline{0.773}  & \underline{0.047} & \underline{0.175}  & \underline{26.66}  & \underline{0.833}  & \underline{0.035}  & \underline{0.155} \\
\rowcolor{color4}
\textbf{LAformer-B}$_D$ &  \textbf{29.53}& \textbf{0.898}& \textbf{0.023}&\textbf{0.125} & \textbf{24.49}& \textbf{0.795}&\textbf{0.044}& \textbf{0.160} & \textbf{26.95} &  \textbf{0.845}& \textbf{0.034}& \textbf{0.143}\\

\bottomrule[0.15em]
\end{tabular}}
\label{table:dpdeblurring}
\end{center}
\end{table*}

\begin{table}[t!]
    \centering
       
    \begin{minipage}[t]{0.50\linewidth} 
        \centering
         \caption{\textbf{Image dehazing} results on synthetic datasets.}
        \setlength{\tabcolsep}{3pt} 
        \resizebox{\linewidth}{!}{ 
            \begin{tabular}{lcccc}
            \toprule[0.15em]
            \multirow{2}{*}{\textbf{Method}} & \multicolumn{2}{c}{\textbf{SOTS-Indoor}~\cite{li2018benchmarking}} & \multicolumn{2}{c}{\textbf{SOTS-Outdoor}~\cite{li2018benchmarking}}  \\ \cmidrule(r){2-5}
             & PSNR $\uparrow$& SSIM $\uparrow$ & PSNR $\uparrow$ & SSIM $\uparrow$  \\ \midrule
            GridDehazeNet~\cite{liu2019griddehazenet} & 32.16 & 0.984 & 30.86 & 0.982 \\
            MSBDN~\cite{dong2020multi} & 33.67 & 0.985 & 33.48 & 0.982\\
            FFA-Net~\cite{qin2020ffa} & 36.39 & 0.989 & 33.57 & 0.984  \\
            AECR-Net~\cite{wu2021contrastive} & 37.17 & 0.990 & N/A & N/A \\
            MAXIM~\cite{tu2022maxim} & 38.11& 0.991& 34.19& 0.985 \\
            DeHamer~\cite{guo2022image} & 36.63 & 0.988 & 35.18 & 0.986  \\
            PMNet~\cite{ye2022perceiving} & 38.41 & 0.990 & 34.74 & 0.985  \\
            FocalNet~\cite{focalnet} & \underline{40.82} & \underline{0.996} & \underline{37.71} & \underline{0.995} \\
            MB-TaylorFormer~\cite{taylorformer} & 40.71 & 0.992 & 37.42 &0.989  \\
            \rowcolor{color4}
            \textbf{LAformer-T} & \textbf{41.17} & \textbf{0.996} &\textbf{38.51} & \textbf{0.995} \\ \bottomrule[0.15em]
            \end{tabular}
        }
        \label{tab:haze_syn}
    \end{minipage}
    \hfill 
    \begin{minipage}[t]{0.47\linewidth} 
        \centering
        \caption{\textbf{Image dehazing} results on real-world datasets.}
        \setlength{\tabcolsep}{3pt} 
        \resizebox{\linewidth}{!}{
            \begin{tabular}{lcccc}
            \toprule[0.15em]
            \multirow{2}{*}{\textbf{Method}} & \multicolumn{2}{c}{\textbf{Outdoor-Haze}~\cite{ancuti2018haze}} & \multicolumn{2}{c}{\textbf{Dense-Haze}~\cite{ancuti2019dense}}  \\ \cmidrule(r){2-5}
             & PSNR $\uparrow$& SSIM $\uparrow$& PSNR $\uparrow$& SSIM $\uparrow$ \\ \midrule
            GridDehazeNet~\cite{liu2019griddehazenet} & 18.92 & 0.672 & 14.96 & 0.536 \\
            MSBDN~\cite{dong2020multi} & 24.36& 0.749& 15.13& 0.555  \\
            FFA-Net~\cite{qin2020ffa} & 22.12& 0.770& 15.70& 0.549   \\
            AECR-Net~\cite{wu2021contrastive} & N/A & N/A& 15.80 &0.470 \\
            Restormer~\cite{restormer} &23.58& 0.768& 15.78& 0.548 \\
            DeHamer~\cite{guo2022image} & \underline{25.11}& 0.777& 16.62& 0.560  \\
            PMNet~\cite{ye2022perceiving} &  24.64& \underline{0.830}& 16.79 & 0.510  \\
            MB-TaylorFormer~\cite{taylorformer} & 25.05 &0.788& 16.66& \underline{0.560}   \\
            AST~\cite{ast} & N/A &N/A& \underline{17.12} &0.550   \\
            \rowcolor{color4}
            \textbf{LAformer-T} & \textbf{25.28} & \textbf{0.932} &\textbf{17.14} & \textbf{0.565} \\ \bottomrule[0.15em]
            \end{tabular}
        }
        \label{tab:haze_real}
    \end{minipage}
\end{table}

\subsection{Image Restoration}

\textbf{Motion Deblurring.} Tab.~\ref{tab:motion} and Tab.~\ref{tab:motion_real} show the motion deblurring results on the synthetic and real-world datasets, respectively. When trained on the synthetic GoPro~\cite{gopro} dataset, our model outperforms all comparison methods on both the GoPro and HIDE~\cite{shen2019human} datasets. When trained on the real RealBlur~\cite{realblur} dataset, our model achieves the best results, surpassing GRL~\cite{grl} by 0.87 dB in PSNR on RealBlur-R. Fig.~\ref{fig:visual_defocus_motion} shows that LAformer produces sharper and clearer results compared to other methods.

We present a comparison of model efficiency on the GoPro dataset in Tab.~\ref{tab:efficiency}, where FLOPs are computed using 256×256 input images. Our method demonstrates superior efficiency compared to other approaches.

\noindent\textbf{Defocus Deblurring.} Tab.~\ref{table:dpdeblurring} shows the comparison results on the DPDD~\cite{dpdd} dataset. Our LAformer consistently outperforms SOTA methods across all three scene types. Specifically, in the combined scenes, LAformer achieves a PSNR improvement of 0.3 dB over the Transformer-based method Restormer~\cite{restormer} in the single-image setting, and a 0.29 dB improvement in the dual-pixel setting. Fig.~\ref{fig:visual_defocus_motion} shows that LAformer restores clear text, whereas other methods result in noticeable blurring or artifacts.

\begin{figure*}[!t]
	
	\centering
	
	\newcommand{\h}{0.105}
	\newcommand{\wa}{0.12}
	\newcommand{\wb}{0.16}
	\newcommand{\g}{-0.7mm}
	\setlength\tabcolsep{1.8pt}
  	\renewcommand{\arraystretch}{1}
	\resizebox{1.00\linewidth}{!} {
			\renewcommand{\h}{0.143}
			\newcommand{\w}{0.22}
				\begin{adjustbox}{valign=t}
					\begin{tabular}{ccccccc}
						\includegraphics[height=\h \textwidth, width=\w \textwidth]{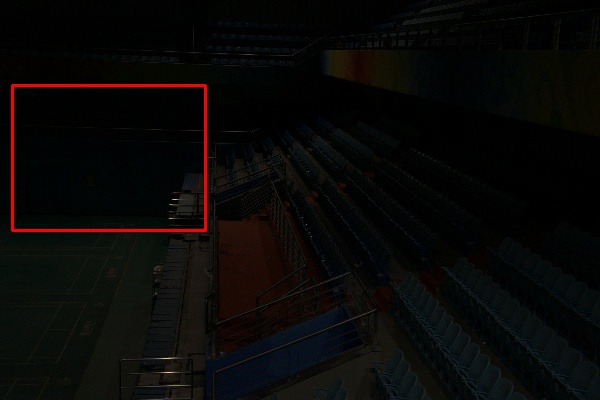} \hspace{\g} &
						\includegraphics[height=\h \textwidth, width=\w \textwidth]{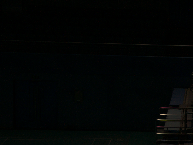} \hspace{\g} &
				\includegraphics[height=\h \textwidth, width=\w \textwidth]{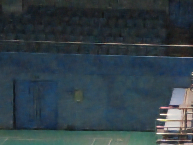} \hspace{\g} &
      			\includegraphics[height=\h \textwidth, width=\w \textwidth]{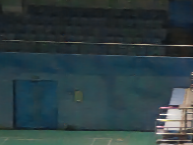} \hspace{\g} &
      			\includegraphics[height=\h \textwidth, width=\w \textwidth]{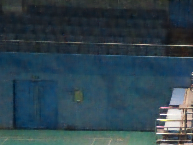} \hspace{\g} &
      			\includegraphics[height=\h \textwidth, width=\w \textwidth]{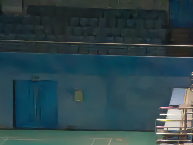} \hspace{\g} &
				\includegraphics[height=\h \textwidth, width=\w \textwidth]{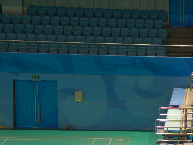} 
						\\
						Low-light Image \hspace{\g} &
						Input \hspace{\g} &
						Restormer~\cite{restormer} \hspace{\g} &
						SNR~\cite{xu2022snr} \hspace{\g} &
						Retinexformer~\cite{cai2023retinexformer} \hspace{\g} &
      					\textbf{LAformer} \hspace{\g} &
                            Target \hspace{\g}
						\\
      					PSNR/SSIM \hspace{\g} &
						6.00/0.188 \hspace{\g} &
						13.63/0.694 \hspace{\g} &
						13.26/0.710 \hspace{\g} &
						13.38/0.707\hspace{\g} &
      					\textbf{13.71}/\textbf{0.740} \hspace{\g} &
                            $+\infty$/1.000 \hspace{\g}
						\\
					\end{tabular}
				\end{adjustbox}
			
	}
   	\renewcommand{\arraystretch}{1}
	\resizebox{1.00\linewidth}{!} {
			\renewcommand{\h}{0.143}
			\newcommand{\w}{0.22}
				\begin{adjustbox}{valign=t}
					\begin{tabular}{ccccccc}
						\includegraphics[height=\h \textwidth, width=\w \textwidth]{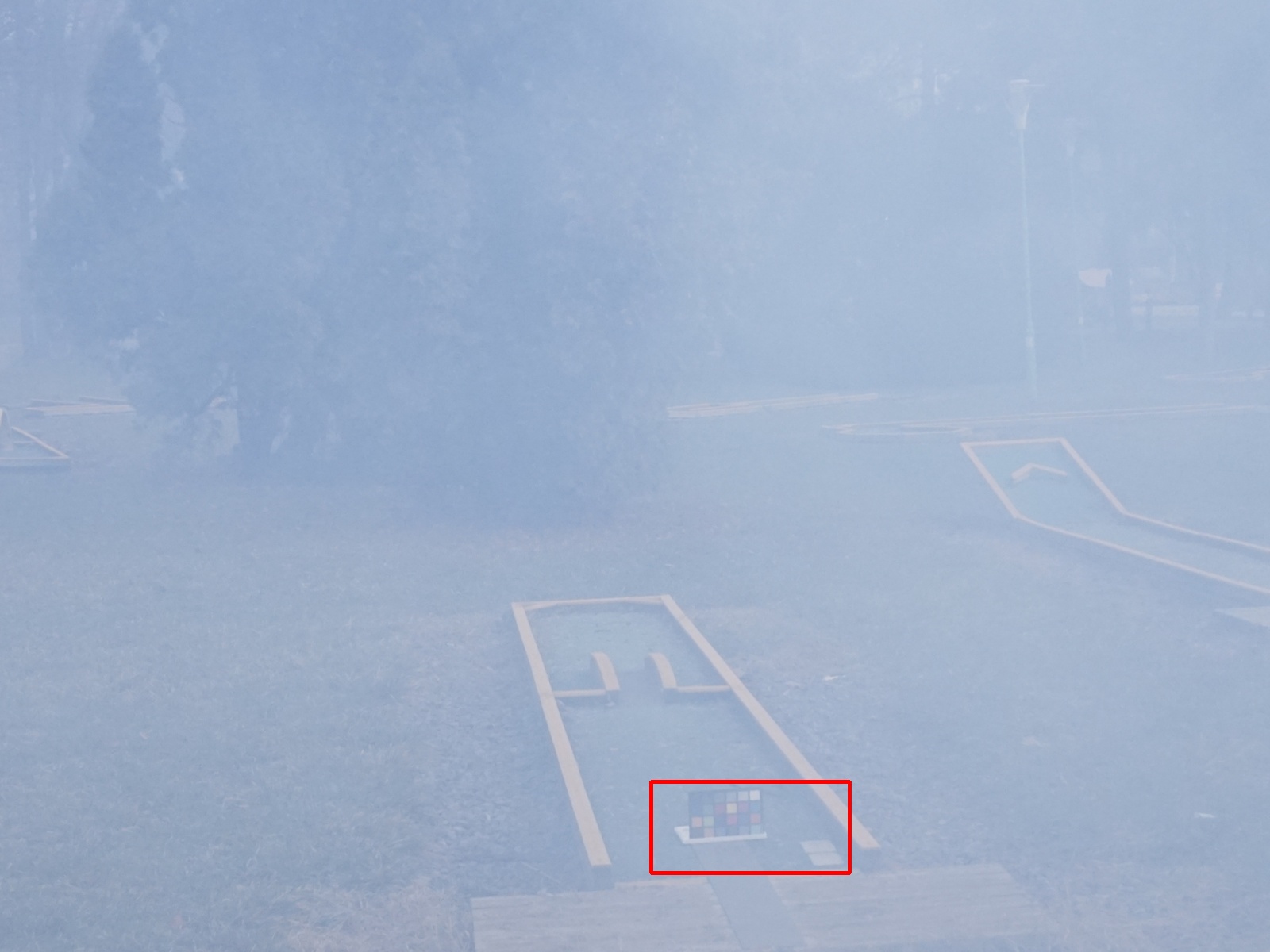} \hspace{\g} &
						\includegraphics[height=\h \textwidth, width=\w \textwidth]{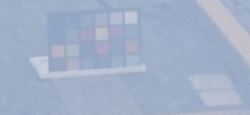} \hspace{\g} &
				\includegraphics[height=\h \textwidth, width=\w \textwidth]{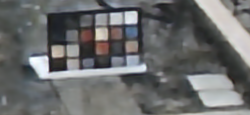} \hspace{\g} &
      			\includegraphics[height=\h \textwidth, width=\w \textwidth]{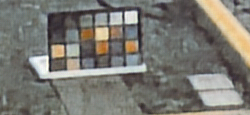} \hspace{\g} &
      			\includegraphics[height=\h \textwidth, width=\w \textwidth]{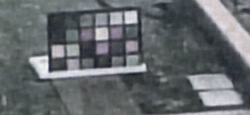} \hspace{\g} &
      			\includegraphics[height=\h \textwidth, width=\w \textwidth]{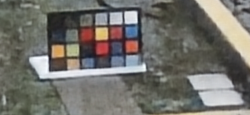} \hspace{\g} &
				\includegraphics[height=\h \textwidth, width=\w \textwidth]{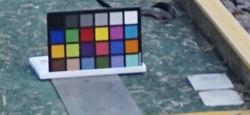} 
						\\
						Hazy Image \hspace{\g} &
						Input \hspace{\g} &
						DeHamer~\cite{guo2022image} \hspace{\g} &
						MB-TaylorFormer~\cite{taylorformer} \hspace{\g} &
						AST~\cite{ast} \hspace{\g} &
      					\textbf{LAformer} \hspace{\g} &
                            Target \hspace{\g}
						\\
      					PSNR/SSIM \hspace{\g} &
						10.53/0.409 \hspace{\g} &
						19.57/0.701 \hspace{\g} &
						19.80/0.707 \hspace{\g} &
						18.18/0.709 \hspace{\g} &
      					\textbf{20.64}/\textbf{0.756} \hspace{\g} &
                            $+\infty$/1.000 \hspace{\g}
						\\
					\end{tabular}
				\end{adjustbox}
			
	}
	\caption{Top row: Visual comparison on the LOL-v1~\cite{wei2018deep} dataset for {\textbf{low-light enhancement}}. Bottom row: Visual comparison on the Dense-Haze~\cite{ancuti2019dense} dataset for {\textbf{image dehazing}}. Zoom in for a better view.}
	\label{fig:visual_haze_lowlight}
\end{figure*}

\noindent\textbf{Image Dehazing.} Tab.~\ref{tab:haze_syn} and Tab.~\ref{tab:haze_real} present the dehazing results on the synthetic and real-world datasets, respectively. LAformer achieves the best performance for all datasets. Fig.~\ref{fig:visual_haze_lowlight} shows that LAformer achieves the dehazing result closest to the ground truth compared to other methods.

\noindent\textbf{Image Desnowing.} Tab.~\ref{tab:snow} shows the desnowing results on three widely used datasets. We follow the dataset split used in FocalNet~\cite{focalnet}. Our method achieves the best performance across three datasets.

\noindent\textbf{Raindrop Removal.} Tab.~\ref{tab:raindrop} shows the raindrop removal results on the Raindrop~\cite{raindrop} dataset. Compared to recent methods AST~\cite{ast} and Histoformer~\cite{histo}, LAformer achieves a PSNR gain of 1.06 dB and 0.32 dB, respectively.

\noindent\textbf{Low-light Enhancement.} As shown in Tab.~\ref{tab:lowlight}, LAformer achieves the best PSNR score for all datasets. Fig.~\ref{fig:visual_haze_lowlight} shows that LAformer produces the clearest visual results, while the results of the comparative methods contain noticeable noise and artifacts.

\noindent\textbf{All-in-One IR.} We further validate the effectiveness of LAformer on the all-in-one IR task. Following the five-task setting of \cite{diffuir,ai2024lora}, we train a single LAformer model to address five different types of degradations. As shown in Tab.~\ref{tab:5task}, our LAformer outperforms DiffUIR across all tasks, delivering significant performance gains. This further highlights the strong representation capacity of LAformer, which achieves superior results without requiring any specialized design for the all-in-one IR task.

\begin{table}[t]
    \centering
    \begin{minipage}[t]{0.48\linewidth}
        \centering
                \caption{\textbf{Raindrop removal} results on the Raindrop~\cite{raindrop} dataset.}
        \setlength{\tabcolsep}{1.5pt}
        \resizebox{\linewidth}{!}{
            \begin{tabular}{l|c|cc|cc}
            \toprule[0.15em]
            \multirow{2}{*}{\textbf{Method}} &\multirow{2}{*}{\textbf{Venue}}&
            \multicolumn{2}{c|}{\textbf{Test-A}} & \multicolumn{2}{c}{\textbf{Test-B}} \\
             &  & PSNR  & SSIM & PSNR & SSIM  \\
            \midrule
            AGAN~\cite{raindrop} & {CVPR'18} & 31.62 & 0.921 & 25.05 & 0.811 \\
            Quan~\etal\cite{quan2019deep} & {ICCV'19} & 31.44 & 0.926 & N/A & N/A \\
            DuRN~\cite{durn} & {CVPR'19} & 31.24& 0.926& 25.32& 0.817 \\
            Restormer~\cite{restormer} & {CVPR'22} & 32.18 & 0.941& N/A & N/A \\
            MAXIM~\cite{tu2022maxim} & {CVPR'22} & 31.87 &0.935& \underline{25.74} & \underline{0.827} \\
            IDT~\cite{idt} & {TPAMI'23} & 31.87 &0.931& N/A & N/A \\
            WeatherDiff~\cite{weatherdiff} & {TAPMI'23} & 32.43 & 0.933 & N/A & N/A \\
            AWRCP~\cite{ye2023adverse} & {ICCV'23}  & 31.93 & 0.931 &N/A&N/A \\
            DTPM~\cite{ye2024learning} & {CVPR'24}  & 32.72 &  0.944 &N/A&N/A \\
            AST~\cite{ast} & {CVPR'24}  & 32.32 &  0.935 &N/A&N/A \\
            Histoformer~\cite{histo} & {ECCV'24}  & 33.06 &{0.944} &N/A&N/A \\
            MOERL~\cite{wang2025moerl} & {ICCV'25}  & \underline{33.21} &\underline{0.945} &N/A&N/A \\
            \rowcolor{color4}
            \textbf{LAformer-S}&-  & \textbf{33.38} & \textbf{0.949} & \textbf{27.27} & \textbf{0.850}  \\
            \bottomrule[0.15em]
            \end{tabular}
        }
        \label{tab:raindrop}
    \end{minipage}
    \hfill 
    \begin{minipage}[t]{0.50\linewidth}
        \centering
        \setlength{\tabcolsep}{1.5pt}
                \caption{\textbf{Low-light enhancement} results on three datasets.}

        \resizebox{\linewidth}{!}{
            \begin{tabular}{lcccccc} 
            \toprule[0.15em]
            \multirow{2}{*}{\textbf{Method}} & \multicolumn{2}{c}{\textbf{LOL-v1}~\cite{wei2018deep}} & \multicolumn{2}{c}{\textbf{LOL-v2-real}~\cite{yang2021sparse}} & \multicolumn{2}{c}{\textbf{LOL-v2-syn}~\cite{yang2021sparse}} \\ 
            \cmidrule(lr){2-3} \cmidrule(lr){4-5} \cmidrule(lr){6-7}
             & {PSNR} & {SSIM} & {PSNR} & {SSIM} & {PSNR} & {SSIM} \\ \midrule
            SID~\cite{chen2019seeing} & 14.35 & 0.436 & 13.24 & 0.442 & 15.04 & 0.610 \\
            IPT~\cite{chen2021pre} & 16.27 & 0.504 & 19.80 & 0.813 & 18.30 & 0.811 \\
            Uformer~\cite{uformer} & 16.36 & 0.771 & 18.82 & 0.771 & 19.66 & 0.871 \\
            Sparse~\cite{yang2021sparse} & 17.20 & 0.640 & 20.06 & 0.816 & 22.05 & 0.905 \\
            RUAS~\cite{liu2021retinex} & 18.23 & 0.720 & 18.37 & 0.723 & 16.55 & 0.652 \\
            SCI~\cite{ma2022toward} & 14.78 & 0.646 & 20.28 & 0.752 & 24.14 & 0.928 \\
            KinD~\cite{zhang2019kindling} & 20.87 & 0.802 & 17.54 & 0.669 & 13.29 & 0.578 \\
            MIRNet~\cite{zamir2020learning} & 24.14 & 0.830 & 20.02 & 0.820 & 21.94 & 0.876 \\
            DRBN~\cite{yang2020fidelity} & 19.86 & 0.834 & 20.13 & 0.830 & 23.22 & 0.827 \\
            URetinex-Net~\cite{wu2022uretinex} & 21.33 & 0.835 & 21.16 & 0.840 & 24.14 & 0.928 \\
            Restormer~\cite{restormer} & 22.43 & 0.823 & 19.94 & 0.827 & 21.41 & 0.830 \\
            MRQ~\cite{liu2023low} & 25.24 & 0.855 & 22.37 & 0.846 & 25.94 & 0.935 \\
            LLFlow~\cite{wang2022low} & 25.19 & \textbf{0.870} & 26.53 & \textbf{0.892} & 26.23 & \underline{0.943} \\
            SNR~\cite{xu2022snr} & 24.61 & 0.842 & 23.92 & 0.849 & 25.77 & 0.894 \\
            Retinexformer~\cite{cai2023retinexformer} & \underline{27.18} & 0.850 & \underline{27.71} & 0.856 & \underline{29.04} & 0.939 \\
            \rowcolor{color4}
            \textbf{LAformer-T} & \textbf{27.28} & \underline{0.868}  &\textbf{27.92} &\underline{0.868} &\textbf{30.28} & \textbf{0.952}\\ \bottomrule[0.15em]
            \end{tabular}
        }
        \label{tab:lowlight}
    \end{minipage}
\end{table}

\begin{table*}[t]
    \centering
    \caption{{\textbf{All-in-one image restoration}} results for five tasks. We strictly follow the setting of DiffUIR~\cite{diffuir}.}
\scalebox{0.66}{
\setlength{\tabcolsep}{3pt}    
        \begin{tabular}{ll|c|cc|cc|cc|cc|cc}
        \toprule[0.15em]
        \multicolumn{2}{l|}{\multirow{2}{*}{\textbf{Method}}} & \multirow{2}{*}{\textbf{Venue}} & 
        \multicolumn{2}{c|}{\textbf{Deraining} $(5sets)$} & \multicolumn{2}{c|}{\textbf{Enhancement}} & \multicolumn{2}{c|}{\textbf{Desnowing} $(2sets)$} & \multicolumn{2}{c|}{\textbf{Dehazing}} & \multicolumn{2}{c}{\textbf{Deblurring}} \\
        \multicolumn{2}{c|}{} & \multicolumn{1}{c|}{} & PSNR $\uparrow$ & SSIM $\uparrow$ & PSNR $\uparrow$ & SSIM $\uparrow$ & PSNR $\uparrow$ & SSIM $\uparrow$ & PSNR $\uparrow$ & SSIM $\uparrow$  & PSNR $\uparrow$ & SSIM $\uparrow$ \\
        \midrule[0.1em]
        \multicolumn{2}{l|}{Restormer~\cite{restormer}} & {CVPR'22}  & 27.10 & 0.843 & 17.63 & 0.542  & 28.61 & 0.876 & 22.79 & 0.706 & 26.36 & 0.814  \\
        \multicolumn{2}{l|}{AirNet~\cite{airnet}} & {CVPR'22}  & 24.87 & 0.773 & 14.83 & 0.767 & 27.63 & 0.860 & 25.47 & 0.923 & 26.92 & 0.811 \\
        \multicolumn{2}{l|}{Painter~\cite{wang2023images}} & {CVPR'23}  & 29.49 & 0.868 & 22.40 & 0.872 & N/A & N/A & N/A & N/A & N/A & N/A \\
        \multicolumn{2}{l|}{IDR~\cite{zhang2023ingredient}} & {CVPR'23} & N/A & N/A & 21.34 & 0.826 & N/A & N/A & 25.24 & 0.943 & 27.87 & 0.846  \\
        \multicolumn{2}{l|}{ProRes~\cite{prores}} & {arXiv'23} & 30.67 & 0.891 & 22.73 & 0.877 & N/A & N/A & N/A & N/A & 27.53 & 0.851 \\
        \multicolumn{2}{l|}{PromptIR~\cite{promptir}} & {NeurIPS'23} & 29.56 & 0.888 & 22.89 & 0.847 & 31.98 & 0.924 & 32.02 & 0.952 & 27.21 & 0.817 \\
        \multicolumn{2}{l|}{DACLIP~\cite{daclip}} &{ICLR'24}  & 28.96 & 0.853 & 24.17 & 0.882 & 30.80 & 0.888 & 31.39 & \underline{0.983} & 25.39 & 0.805 \\
        \multicolumn{2}{l|}{DiffUIR~\cite{diffuir}}& {CVPR'24} & \underline{31.03} & \underline{0.904} & \underline{25.12} & \underline{0.907} & 32.65 & 0.927 & \underline{32.94} & 0.956 & 29.17 & 0.864 \\
        \multicolumn{2}{l|}{Varformer~\cite{wang2025navigating}}& {CVPR'25} & 31.33 & 0.913  & 25.13 & 0.917 & N/A & N/A & 32.96 & 0.956 & N/A & N/A \\

\rowcolor{color4}
        \multicolumn{2}{l|}{\textbf{LAformer-B}}&  -  & \textbf{31.55} & \textbf{0.914} & \textbf{26.81} & \textbf{0.924} & \textbf{34.24} & \textbf{0.942} &\textbf{34.28} & \textbf{0.984} & \textbf{30.92} & \textbf{0.910}  \\
        \bottomrule[0.15em]
    \end{tabular}{}
    }
    \label{tab:5task}
\end{table*}

\begin{table}[!t]
  \footnotesize
  \centering
  \caption{ \textbf{Diffusion-based image generation} results for 400K training steps on class-conditional ImageNet 256$\times$256. We use \textit{the same training hyperparameters} as DiT, DiC, and DiG.}
\resizebox{.9\linewidth}{!}{
\setlength{\tabcolsep}{5pt}    
\begin{tabular}{l|c|cc|cc|cc}
\toprule[0.15em]
Method & Venue & Params & FLOPs & FID$\downarrow$  & IS$\uparrow$    & Prec.$\uparrow$ & Rec.$\uparrow$ \\
\midrule
DiT-S/2~\cite{dit} & ICCV'23 & 33M  & 6.06G          & 68.40    & -      & -          & -       \\
DiC-S~\cite{tian2024dic} & CVPR'25  & 33M & 5.9G            & 58.68  & 25.82 & - & - \\

DiG-S/2~\cite{zhu2024dig}& CVPR'25 & 33M & 4.30G & 62.06 & 22.81 & 0.39 & 0.56 \\
\rowcolor{blue!6}
\textbf{LAformer} &-   & 33M & 4.32G          & \textbf{50.39}  &  \textbf{28.64} & \textbf{0.44} & \textbf{0.59} \\

\midrule
DiT-B/2 & ICCV'23 & 130M & 23.02G         & 43.47    & -      & -          & -       \\
DiC-B  & CVPR'25 & 130M & 23.5G          & 32.33 & 48.72 & - & - \\
DiG-B/2 & CVPR'25   & 130M & 17.07G        & 39.50 & 37.21 & 0.51 & 0.63 \\
\rowcolor{blue!6}
\textbf{LAformer} &-  & 131M &   17.32G       & \textbf{26.92} &  \textbf{53.17} &  \textbf{0.56} & \textbf{0.64} \\
\midrule
DiT-L/2& ICCV'23  & 458M & 80.73G          & 23.33    & -      & -          & -       \\
DiG-L/2 & CVPR'25    & 458M & 61.66G      & 22.90  & 59.87 & 0.60 & 0.64 \\
\rowcolor{blue!6}
\textbf{LAformer}&- & 463M & 60.98G         & \textbf{13.42}  & \textbf{85.52} &  \textbf{0.66} & \textbf{0.65} \\
\midrule
DiT-XL/2& ICCV'23 & 675M & 118.66G       & 19.47     & -      & -          & -       \\
DiC-XL& CVPR'25  & 702M & 116.1G          & 13.11  & \textbf{100.15} & - & - \\
DiG-XL/2& CVPR'25    & 676M & 89.40G         & 18.53  & 68.53 & 0.63 & 0.64 \\
\rowcolor{blue!6}
\textbf{LAformer}&- & 682M  & 89.91G         & \textbf{11.31} & 98.21 &  \textbf{0.67} & \textbf{0.66} \\

\bottomrule[0.15em]
\end{tabular}}
  \label{tab:in256}
\end{table}

\begin{figure*}[t]
    \centering
    \includegraphics[width=1.0\linewidth]{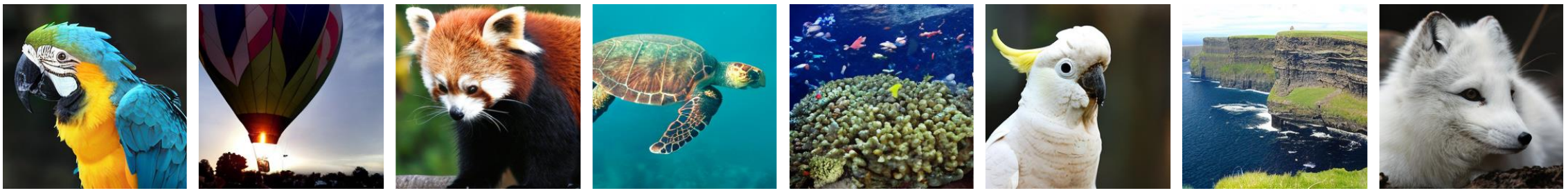}
    \caption{\textbf{Generated samples from LAformer} trained on the class-conditional ImageNet dataset at 256$\times$256 resolution.}
    \label{fig:laformer_gen}
\end{figure*}

\begin{table}[t]
  \footnotesize
  \centering
  \caption{ \textbf{Flow-based image generation} results for 400K steps. We use \textit{the same training hyperparameters} as SiT and FlowDCN.}
\resizebox{0.9\linewidth}{!}{
\setlength{\tabcolsep}{5pt}    
\begin{tabular}{l|c|cc|cc|cc}
\toprule[0.15em]
Method& Venue & Params & FLOPs & FID$\downarrow$  & IS$\uparrow$    & Prec$\uparrow$ & Rec$\uparrow$ \\
\midrule
SiT-S/2~\cite{ma2024sit}&ECCV'24 & 33M  & 6.06G         & 57.64    & 24.78      & 0.41         & 0.60       \\
FlowDCN-S/2~\cite{wang2024flowdcn} &NeurIPS'24 & 30.3M  & 4.36G         & 54.6    & 26.4      & -         & -       \\
\rowcolor{blue!6}
\textbf{LAformer}&-   & 33M & 4.32G         & \textbf{46.92}  &  \textbf{31.74} & \textbf{0.46} & \textbf{0.61} \\

\midrule
SiT-B/2&ECCV'24  & 130M & 23.02G         & 33.02    & 43.71      & 0.53          & 0.63       \\
FlowDCN-B/2 &NeurIPS'24 & 130M  &17.87G         & 28.5    & 51      & -         & -       \\
\rowcolor{blue!6}
\textbf{LAformer}&-  & 131M &   17.32G       & \textbf{24.37} &  \textbf{55.53} &  \textbf{0.57} & \textbf{0.64} \\
\midrule
SiT-L/2&ECCV'24 & 458M & 80.73G          & 18.79    & 72.02      & 0.64          & 0.64       \\
FlowDCN-L/2 &NeurIPS'24 & 421M  &63.51G         & 13.8    & 85      & -         & -       \\
\rowcolor{blue!6}
\textbf{LAformer}&- & 463M & 60.98G         & \textbf{12.38}  & \textbf{87.34} &  \textbf{0.66} & \textbf{0.65} \\
\midrule
SiT-XL/2&ECCV'24 & 675M & 118.66G       & 17.19     & 76.52      & 0.65          & 0.63       \\
FlowDCN-XL/2 &NeurIPS'24 & 618M  &93.24G         & 11.3    & 97      & -         & -       \\
\rowcolor{blue!6}
\textbf{LAformer}&- & 682M  & 89.91G         & \textbf{10.98} & \textbf{102.04} &  \textbf{0.67} & \textbf{0.66} \\

\bottomrule[0.15em]
\end{tabular}}
  \label{tab:in256_flow}

\end{table}

\begin{table}[!t]
\captionsetup{font=small}%
\scriptsize
\center
    \caption{Ablations on GoPro~\cite{gopro} for RELA designs.}
\begin{center}
\resizebox{0.7\linewidth}{!}{
\setlength{\tabcolsep}{3pt}    
\begin{tabular}{l|cc|c}
    \toprule
        RELA  & Params (M) & FLOPs (G) & PSNR (dB)
         \\ \midrule
         $\psi(\cdot)=\mathrm{ReLU}(\cdot)$  & 24.89 & 144.33 & 33.38 \\
         \rowcolor{color4}
         $\psi(\cdot)=1+\mathrm{ELU}(\cdot)$  & 24.89 & 144.33 & 33.40 \\
         \midrule
         3$\times$3 DWC  & 24.79 & 143.11 & 33.35 \\
         \rowcolor{color4}
         \textbf{5$\times$5 DWC}  & 24.89 & 144.33 & 33.40 \\
         7$\times$7 DWC  & 25.03 & 146.16 & 33.41 \\
         \bottomrule
    \end{tabular}}
\end{center}
        \label{tab:ablation_rela}
\end{table}

\begin{table}[t]
\captionsetup{font=small}%
\scriptsize
\center
\caption{Ablations on SOTS-Indoor~\cite{li2018benchmarking} for LAformer-T.}
\begin{center}
\resizebox{0.7\linewidth}{!}{
\setlength{\tabcolsep}{3pt}    
\begin{tabular}{l|cc|c}
    \toprule
        Method  & Params (M)  & FLOPs (G)  & PSNR (dB)          \\ \midrule
         Vanilla LA~\cite{katharopoulos2020transformers}  & 6.18 & 29.08 & 39.28 \\
         Window SA~\cite{swinir}  & 6.20 & 30.19 & 40.76 \\
         Transposed SA~\cite{restormer}  & 6.26 & 29.87 & 40.52 \\
         \rowcolor{color4}
         \textbf{Rank Enhanced LA}  & 6.25 & 29.84 & 41.17 \\
         \midrule
         w/o CAB  & 6.13 & 30.79 & 40.81 \\
        \rowcolor{color4}
         \textbf{Dual Attention}  & 6.25 & 29.84 & 41.17 \\
         \midrule
        MLP  & 6.59 & 30.96 & 40.20 \\
        CG-FFN w/o DWC  & 6.24 & 30.16 & 40.11 \\
        CG-FFN w/o GLU  & 6.28 & 29.60 & 40.86 \\
        \rowcolor{color4}
        \textbf{CG-FFN}  & 6.25 & 29.84 & 41.17 \\
         \bottomrule
    \end{tabular}}
\end{center}
    \label{tab:ablation_model}
\end{table}

\subsection{Visual Generation}
Following \cite{dit,ma2024sit,ai2026dico}, we conduct class-conditional image generation experiments on ImageNet~\cite{deng2009imagenet} at 256$\times$256 resolution. We adjust the depth and width of LAformer to match the parameter count and computational cost of baselines. \textbf{\textit{We follow exactly the same experimental setup as DiT and SiT to ensure a fair comparison.}} Tab.~\ref{tab:in256} shows that across models of varying scales (ranging from approximately 33M to 680M parameters), our LAformer achieves superior generation performance with lower computational cost compared to DiT. This demonstrates the efficiency and scalability of LAformer in both diffusion-based and flow-based visual generation. Fig.~\ref{fig:laformer_gen} shows that our method is capable of generating high-quality images efficiently.

\subsection{Ablation Study}
\label{sec:abla}
As shown in Tab.~\ref{tab:ablation_model} and Tab.~\ref{tab:ablation_rela}, we perform detailed ablation studies on SOTS-Indoor~\cite{li2018benchmarking} and GoPro~\cite{gopro} datasets to analyze the role of each component in LAformer. For fair comparisons, all models are designed with comparable complexity. FLOPs are computed on 256 $\times$ 256 images. More ablations and analysis are provided in the \textbf{Appendix}.

\noindent\textbf{Rank Enhanced Linear Attention.} As shown in Tab.~\ref{tab:ablation_model},  RELA significantly improves performance over vanilla linear attention, achieving a 1.89 dB PSNR gain on SOTS-Indoor with only a modest increase of 0.07M parameters and 0.76G FLOPs. We also compare RELA with the commonly used window self-attention~\cite{swinir} and transposed self-attention~\cite{restormer} in image restoration. With similar parameter counts and FLOPs, our RELA achieves a PSNR improvement of 0.41 dB and 0.65 dB, respectively. This demonstrates the global perception capability of RELA, which contributes to improved restoration performance.

We also conduct ablations on the design of RELA, as shown in Tab.~\ref{tab:ablation_rela}. We analyze the choice of activation function $\psi(\cdot)$, with experimental results indicating minimal impact on PSNR performance. This suggests that the effectiveness of our RELA is not dependent on meticulously chosen activation functions. We further analyze the impact of the depth-wise convolution (DWC) kernel size on the results. Considering both computational efficiency and performance, we ultimately select a $5 \times 5$ kernel.

\noindent\textbf{Channel Attention Block.} Tab.~\ref{tab:ablation_model} shows that removing CAB results in a 0.36 dB drop in PSNR, highlighting the importance of channel-wise global information for IR.

\noindent\textbf{Convolutional Gated FFN.} Tab.~\ref{tab:ablation_model} indicates that the DWC in CG-FFN significantly enhances LAformer's ability to extract local information, markedly improving model performance. Removing it results in nearly a 1 dB drop in PSNR.

\section{Conclusion}
In this paper, we introduce linear attention to image restoration, aiming to overcome the complexity limitations of Transformer. We uncover the inherent low-rank limitations of vanilla linear attention and propose a simple yet effective solution: Rank Enhanced Linear Attention (RELA). RELA enhances the rank of features through a lightweight depth-wise convolution, thereby improving its representation capability. Building upon RELA, we introduce LAformer, a Vision Transformer designed to achieve effective global perception through the integration of linear attention and channel attention. Comprehensive experiments show that LAformer surpasses SOTA IR methods in performance, while also offering considerable computational efficiency gains. Furthermore, we extend LAformer to visual generation, showcasing its excellent efficiency and scalability. We aim to further scale up LAformer and extend it to broader generative tasks, including text-to-image generation.

%
%
\bibliographystyle{splncs04}
\bibliography{main}
\end{document}